\documentclass[11pt]{article}
\usepackage{acl2014}
\usepackage{times}
\usepackage{url}
\usepackage{latexsym}
\usepackage[utf8]{inputenc}
\usepackage{amssymb,amsmath,amsthm}
\usepackage{fancyhdr}
\usepackage{graphics}
\usepackage{graphicx}
\usepackage{caption}
\usepackage{subfig}
\usepackage{setspace}
\usepackage{cite}
\usepackage{bbding}

\title{Evaluating Bregman Divergences for Probability Learning from Crowd}

\author{Francisco Mena \\
  Univ. T\'ecnica Federico Santa Mar\'ia\thanks{Contact to francisco.mena.13@sansano.usm.cl} \\
  Depto de Inform\'atica \\
  Santiago, Chile \\ \And
  Ricardo \~Nanculef \\
  Univ. T\'ecnica Federico Santa Mar\'ia \\
  Depto de Inform\'atica \\
  Santiago, Chile \\
}
\date{}

\begin{document}

%%%%%%%%PORTADA%%%%%
\maketitle            

\begin{abstract}
The crowdsourcing scenarios are a good example of having a probability distribution over some categories showing what the people in a global perspective thinks. Learn a predictive model of this probability distribution can be of much more valuable that learn only a discriminative model that gives the most likely category of the data. Here we present differents models that adapts having probability distribution as target to train a machine learning model. We focus on the Bregman divergences framework to used as objective function to minimize. The results show that special care must be taken when build a objective function and consider a equal optimization on neural network in Keras framework.
\end{abstract}

\section{Introduction}

Know the probability distribution over different categories (Discrete variables) on some domain can be very valuable when one want a predictive model that can give some prediction together with the uncertain of it. The classic machine learning models try to learn a discriminative model over only the truth category of the data. 

When face a problem of learn all the probability distribution over the categories of some data, the classic machine learning method  may not fit well. This is a different objective, because the model need to learn to give a prediction even on the least likely categories and with this give the uncertain, i.e. avoid to assign a \textit{priori} zero probability to the categories that the data does not belong.

The crowdsourcing platform, such as Amazom Mechanical Turk (AMT)\footnote{http://www.mturk.com}, allows one to obtain various annotations over some dataset, with multiple annotations (not the same) per data. With this one can group all the annotations in one vector of \textit{repeats} and then normalize to get probabilities of each category representing what a common/regular annotator thinks or behave. Then, in this scenario a problem with two categories $(dog, cat)$, an image of a big cat can have probability distribution of $(0.8, 0.2)$ that show how an annotator behave, also how she can get confused and give a wrong annotation over the image.
The experimental work of \cite{snow2008cheap} over different text datasets, shows that multiple inexpert annotators can perform similar to expert annotators, having a strong correlation between them. So we can assumed that multiple annotators can have a standard good behave.

In this application we need to measure some function among two probability distribution as a cost function to optimize, the commonly used in the state of the art  is the KL divergence \cite{thomas1991elements}, that measure the difference between two pdfs. Here we explore different dissimilarity measures between vectors (probabilities among different categories). We focus on the domain of the Bregman divergences \cite{bregman1967relaxation} that measure dissimilarity between objects and by itself is not a metric.

We measure different functions trying to understand what kind of objective functions work better. The results report that Bregman divergences and similar objectives function turned out not behave in the same way. We suspect that this is because the approximate optimization that does the neural network framework, as the stochastic optimization or the approximate functional derivatives.

The paper structure is as follow: In Section 2 we formally we define the problem and what we are facing. In the next section (3) we present the models proposed to solved the problem and compare between them, while the  evaluation metrics to compare is in Section 4. Section 5 present the Related work and in Section 6 we show the results of all the experimentation. Finally the conclusion are shown in Section 7.

\section{Problem}

The task is given a dataset of $N$ pairs $\{(x_i,p_i)\}_{i=1}^N$, with $x_i \in \mathbb{R}^d$ the data and $p_i \in \mathbb{R}^K$ the vector of \textit{repeats} normalized, aka a vector of the probabilities of each category $K$ on the data,  we need to learn a model that maps the data to the probabilities $f: \mathbb{R}^d \rightarrow \mathbb{R}^K, f(x_i) = \hat{p}_i $.

This is a different objective that the classic machine learning that trying to learn only the most probable category ($argmax$). Here we have a loss function that use all the probability distribution to learn $\ell (p_i,f(x_i))$ and jointly learn correctly the uncertainty of the predictions. 

To build the data probabilities, in where some cases come with the data, we assumed that the data and the annotators model correctly the probability of each category, aka ground truth. With $r_{ij}$ the vector of \textit{repeats} that store the number of times that the data $i$ was annotated by category $j$.
\begin{equation}
    p_{ij} = \frac{r_{ij}}{\sum_l r_{il}}
\end{equation}

\section{Models}

The proposed models trying to solve the problem are the well studied deep neural network functions \cite{lecun2015deep} to model $f$ with different objective functions that adapts probabilities. Particularly here we work with the Bregman divergences \cite{bregman1967relaxation}. Here we define the objective function to compare, which are evaluated on every pair of examples in a \textit{batch} and then merge together with an arithmetic average, as a standard neural network.

\subsection{Based on Keras}
Firstly we define some common used in Keras \cite{chollet2015keras} metrics to evaluate deep learning models.

\begin{itemize}
\item \textit{Mean Squared Error (MSE)}:
\begin{equation}
MSE(p_i,\hat{p}_i) = \frac{1}{K} \sum_j^K  (p_{ij} - \hat{p}_{ij})^2
\end{equation}

\item \textit{Root Mean Squared Error (RMSE)}:
\begin{equation}
RMSE(p_i,\hat{p}_i) = \sqrt[]{ \frac{1}{K} \sum_j^K ( p_{ij} - \hat{p}_{ij} )^2 }
\end{equation}
\item \textit{Cross-entropy}
\begin{equation}
H(p_i,\hat{p}_i) =  \sum_j^K - p_{ij} \log \hat{p}_{ij}
\end{equation}
\item \textit{Reverse KL}:
\begin{equation}
KL(\hat{p}_{i}||p_i) = \sum_j^K \hat{p}_{ij} \log \frac{\hat{p}_{ij}}{p_{ij}}
\end{equation}

\item \textit{Jensen Shanon divergence} \cite{lin1991divergence} (also known as \textit{symmetric KL}):
\begin{equation}
JS(p_i,\hat{p}_i) = \frac{1}{2} \sum_j^K  KL(p_{ij} || m_{ij} )  +   KL(\hat{p}_{ij} || m_{ij} )
\end{equation}
With $m_{ij} = \frac{p_{ij}+\hat{p}_{ij}}{2}$

\end{itemize}

\subsection{Bregman divergences}

Here we define the divergence (inverse to similarity) of Bregman \cite{chen2008metrics,banerjee2005clustering} to measure the difference between two probability distribution. These functions come from a family that share some properties, due they are derived from a general framework/structure.

Given $\Phi$, a strictly convex differentiable function , the Bregman divergence $d_ {\Phi}$ is define as:
\begin{equation}
d_{\Phi}(x,y) = \Phi(x) - \Phi(y) - \langle x-y, \nabla (y) \rangle
\end{equation}

With $\langle a , b \rangle$ the \textit{inner product} between $a$ and $b$. As can be seen the order that is given to $d_{\Phi}$ matters, so as it does not fulfill the symmetry or triangular inequality properties, is not a defined as a metric. Nonetheless, it has some other properties that are good for optimization purpose:
\begin{itemize}
\item Convex: on his first argument $x$.
\item Non-negative: $d_{\Phi}(x,y) \geq 0$ for every $x, y$.
\item Duality: If $\Phi$ has a convex conjugate can be used.
\item The median as a minimum in random scenario: Given a set of random vectors, the minimum of $d_{\Phi}(x,y)$ for $y$, given any function $\Phi$ and $x$, is the median of the vectors \cite{banerjee2005clustering}.
\end{itemize}

Then, the divergence $d_{\Phi}(p_i,\hat{p}_i)$ with different $\Phi$ functions:

\begin{itemize}
\item \textit{Square Euclidean Distance} (also known as \textit{sum of square error/sse)}, for $\Phi(p_i) = || p_i ||^2$:
\begin{equation}
SSE(p_i,\hat{p}_i) = \sum_j^K  (p_{ij} - \hat{p}_{ij})^2
\end{equation}

\item \textit{Forward KL}, for negative entropy function, $\Phi(p_i) = \sum_j^K p_{ij} \log p_{ij} $
\begin{equation}
KL(p_i \mid \mid \hat{p}_i) = \sum_j^K p_{ij} \log \frac{p_{ij}}{\hat{p}_{ij}}
\end{equation}

\item \textit{Generalized I divergence}, similar to \textit{Forward KL} but generalized to the positive reals\footnote{\textit{Forward KL} is for a domain of discrete values}.
\begin{equation}
GenI(p_i \mid \mid \hat{p}_i) = \sum_j^K p_{ij} \log \frac{p_{ij}}{\hat{p}_{ij}}  - (p_{ij}-\hat{p}_{ij})
\end{equation}

\item \textit{Itakura Saito distance}, for $\Phi(p_i) = - \log p_{i} $:
\begin{equation}
IS(p_i,\hat{p}_i) =  \sum_j^K \frac{p_{ij}}{\hat{p}_{ij}} - \log \frac{p_{ij}}{\hat{p}_{ij}}  - 1
\end{equation}
\end{itemize}

We hope that a evaluating function (objective function) based on probabilities achieved a best behavior that a standard evaluation function for continuous variables.

\section{Metrics}

In order to fair comparison between the effect of different objective functions, we use some normalized metrics across the different evaluation function used in training:
\begin{itemize}
\item Convergence delta
\begin{equation} \label{deltaconvergencia}
\Delta(t) = \frac{\mid loss^{(t)} - loss^{(t+1)} \mid}{loss^{(t)}}
\end{equation}
With $t$ the instant during training, analogous to epochs.
%\item Top $k$ accuracy: the trained model gives the $k$ most likely categories and it gets check if between these $k$ is the true category ($argmax_k \ \ p_{ik} $), Top-1 is the hard majority voting on the prediction.
\item \textit{Macro} F1 score between the category with high probability:
\begin{equation} \label{f1score}
F_1^M = \frac{1}{K} \sum_{j}^K  2 \frac{P_j\cdot R_j}{P_j+R_j}
\end{equation}
With $P_j$ and $R_j$ the precision and recall over category $j$, respectively.

\item \textit{Normalized discounted cumulative gain} (NDCG), a metric from \textit{learning to rank} \cite{cao2007learning}, with the objective to measure the order of predicted probabilities.

\item \textit{Accuracy on Ranking Decrease}: 
\begin{equation} \label{accuracyranking}
acc_{rank} = \frac{1}{N} \sum_i^N \frac{1}{K} \sum_k^K \cfrac{I(y_i^{(k)} = \hat{y}_i^{(k)} ) } {k}
\end{equation}
With $y_i^{(k)}$ the real category of data $i$ in position $k$ and $I$ the indicator function.
\end{itemize}

\section{Related Work}

Since the Bregman divergences was proposed \cite{bregman1967relaxation}, several works recently has studied the benefits of this divergences. For instances, \cite{vemuri2011total} propose a different way to find the optimum or \textit{t-center} for the objective, as find the representative in a cluster algorithm, for example for \textit{MSE} the center is the mean. Here he present a robust and efficient formulate to seek a center of a different formulation of Bregman divergences.

The work of \cite{banerjee2005clustering} use the Bregman divergence to measure data distance and cluster data. This work is closed related because he used the Bregman divergences as objective functions as us but on unsupervised scenario. The good results shown here say that the Bregman divergences can be powerful on recognize pattern and is a good dissimilarity measure to cluster data. Some Bregman divergences has been applied in order to train a GAN (Generative Adversarial Network) \cite{nowozin2016f}, also another unsupervised scenario where it shows that train with divergences can be done and get some advantages.

 Another application of the Bregman divergence is the one of \cite{sugiyama2012density}, in which porpose a new efficient way to estimate the ratio of probability densities through the framework of Bregman.

Since the Bregman divergences has shown as a strong framework of dissimilarity measure we focus on this framework to base our different objective functions.

\section{Experiments}

\begin{figure*}[t]
  \centering
    \includegraphics[width=0.8\textwidth]{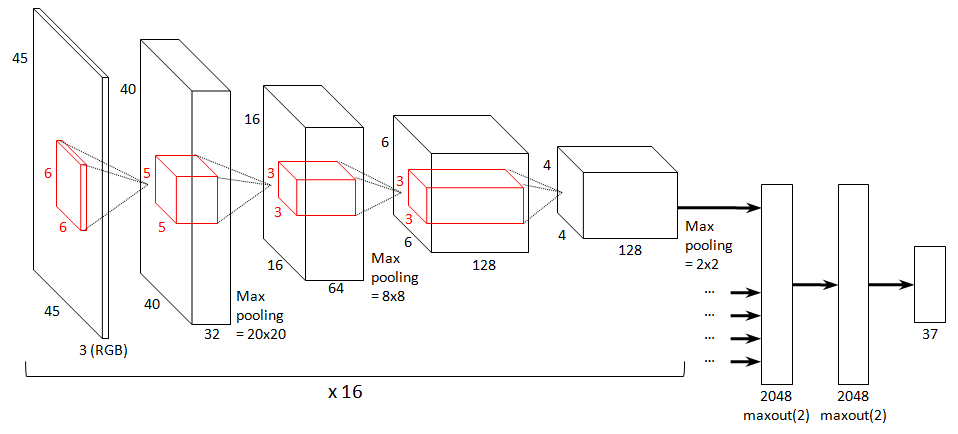}
    \caption{Convolutional network on the GalaxyZoo challenge from Kaggle\footnote{benanne.github.io/2014/04/05/galaxy-zoo.html}.}
    \label{arquitectura}
\end{figure*}

The experiments was realized with deep neural network, trained with a GPU, GeForce GTX 1060 (6 GB) and we repeat the experiment 4 times to normalize the random initialize and optimization of the neural network. we set Adam optimizer \cite{kingma2014adam} with the Glorot initialize of weights \cite{glorot2010understanding}. The batch size is set to 128 and a limit of 20 epochs. We used standard training and test set split of (70/30)\% respectively.

\subsection{Data} 

The first data used is an image data known as GalaxyZoo\footnote{www.galaxyzoo.org}. This project start by astronomer of the Oxford University in 2007 \cite{lintott2008galaxy} in where ask people in thoroughfare that classify their dataset of million of galaxies (thanks to SDSS\footnote{www.sdss.org}). The worked dataset is a small subset of this with he probabilities about different morphologies of the galaxy through thousand of volunteers (annotators).

We work with the Kaggle\footnote{www.kaggle.com} dataset of this, which are 60 thousand RGB images which we re-size to 100x100 pixels. The categories also are a subset of all the answers/annotations, which correspond to 7 answer to question about the galaxy morphology.

\begin{enumerate}
\item How round is the smooth of the galaxy?
\begin{enumerate}
\item Completely round
\item between
\item Cigar shaped
\end{enumerate}
\item What type of disk is the galaxy?
\begin{enumerate}
\item A view edge-on disk
\item Spiral tight
\item Spiral medium
\item Spiral loose
\item Normal disk
\end{enumerate}
\item Is it a Galaxy?
\begin{enumerate}
\item Is a Star or artifact
\end{enumerate}
\end{enumerate}

The categories are mutually exclusive so only one can be given. The one with higher probability over all the dataset are \textit{Normal disk}  and \textit{Smooth between}.\\

The second data what we used is a text data also provided by Kaggle platform, the Stock tweets emotion\footnote{www.kaggle.com/fernandojvdasilva/stock-tweets-ptbr-emotions}. Here we also has multiple annotations by every tweet (wrote in Portuguese) about the emotion express in there. 

The 9 categories represent the emotion of the tweet: \textit{joy}, \textit{sadness}, \textit{trust}, \textit{disgust}, \textit{surprise}, \textit{anticipation}, \textit{anger}, \textit{fear} and \textit{neutral}, where this last one is the category with higher probability over the dataset.

\subsubsection{Architectures}

The model to work and process the images is similar to the presented by the winner of the competition and presented in Figure \ref{arquitectura}. Is a convolutional model \cite{lecun1995convolutional} of 3 convolutional blocks, $C\rightarrow P$, with $P$ the max pooling layer of pool size 2 and $C$ the convolution of kernel size 3 and number of filters 32, 64 and 128 respectively. This is followed by two dense layers with 512 units and activation function ReLU for all.

The model that process the text data is a standard recurrent neural network of two layers with Gated Recurrent Unit (GRU) \cite{chung2014empirical} as gates.

In both models there is a final dense layer with \textit{softmax} activation that gives the predictive probabilities over the categories.

\subsection{Results and Discussion}

\begin{figure}[t]
    \centering
    \subfloat[Objective function limit = 1]{{\includegraphics[width=0.5\textwidth]{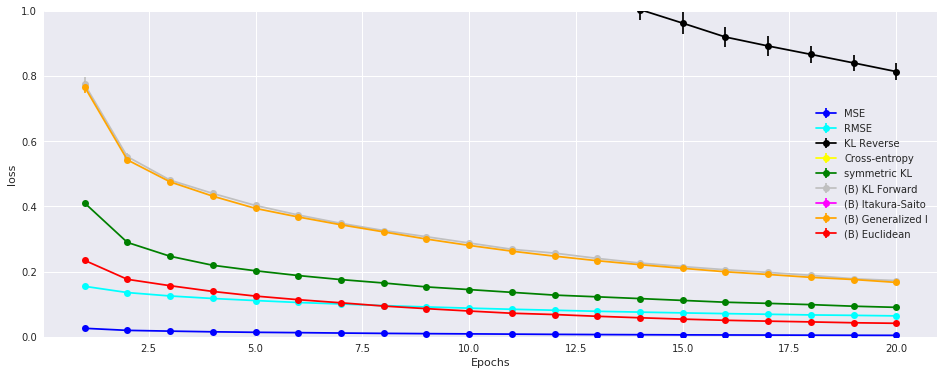} }}
    \qquad
    \subfloat[Objective function limit = 0.4]{{\includegraphics[width=0.5\textwidth]{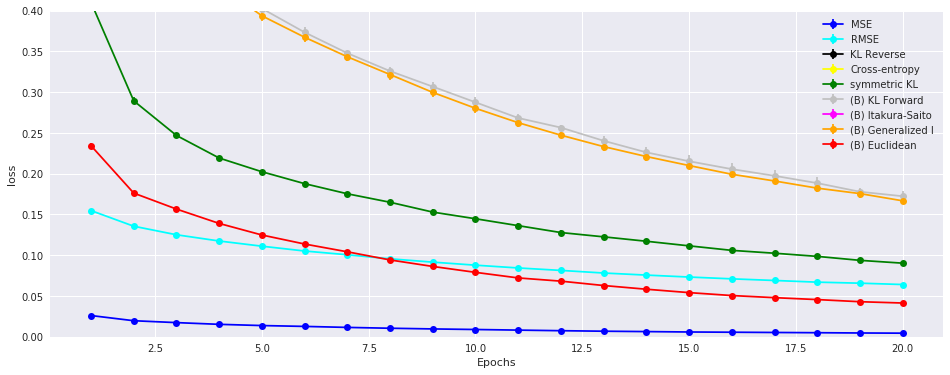} }}
    \caption{Progress of used objective function trough nomber of epochs}
    \label{lossevolution}%
\end{figure}

As a prior analysis we show the result and compare the effect of the different objective function over the GalaxyZoo dataset. We present the general behavior of the loss function in Figure \ref{lossevolution}. Some comments about this is that the scale of the Itakura Saito distance is much higher that the rest (magnitude close to 50), while the numeric metrics from regression as \textit{MSE}, \textit{RMSE} and Euclidean has a lower domain, less than $0.3$. Another observation is that the behavior of Forward KL is practically the same to Generalized I, also Cross Entropy and Jensen-Shannon have similar curvature in the progress of objective function. On the other hand, the curvature of the loss function \textit{MSE} is similar to \textit{RMSE}. This is produced due the similitude of the objective function, because the differences are some multiplicative constant.

\begin{figure}[t]
    \centering
    \subfloat[During all epochs ]{\includegraphics[width=0.5\textwidth]{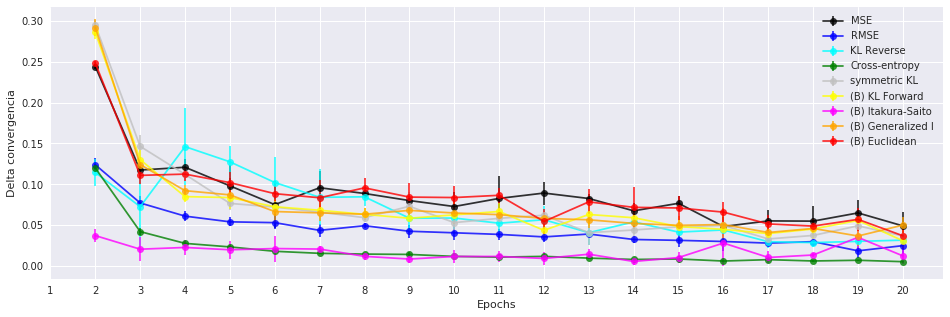} }
    \qquad
    \subfloat[Stopped if less than 0.05]{{\includegraphics[width=0.5\textwidth]{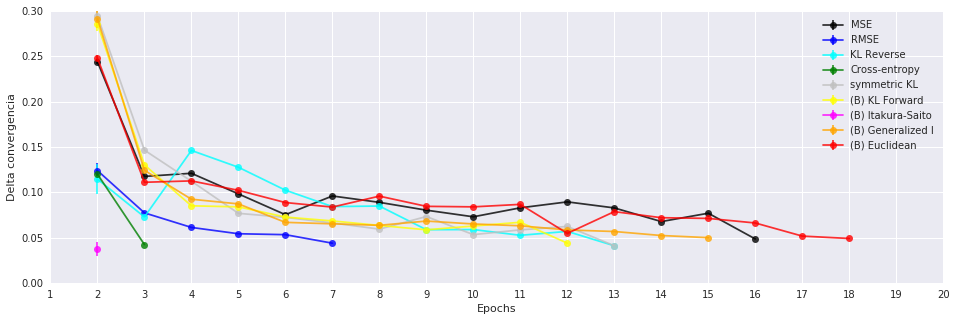} }}
    \caption{Progress of delta convergence trough number of epochs}
    \label{deltaloss}%
\end{figure}

We show a fair comparison of the different scale of objective function in the delta convergence (Equation \ref{deltaconvergencia}) in Figure \ref{deltaloss}. It can be seen that each objective function converge in a way, for example Itakura Saito distance is the first in stop the variation (fast convergence). In the second place it is the Cross Entropy, converging in epoch 3, followed by \textit{RMSE} in epoch 7. The last in converge with a limit set to $0.05$ of variation, after epoch 15, are Generalized I, \textit{MSE} and Euclidean. Here is shown that no pattern can be found between probabilities loss function and numeric continuous loss function except between the last in converge, because its functions share the subtract between the real and the predicted values $(p-\hat{p})$. In this cases the derivative becomes proportional directly to the model and may cause that the model keeps learning on those epochs.

The Table \ref{macrof1} shows the result of the macro F1 metric in both sets of data. It can bee seen that the worst result correspond to the Itakura Saito distance, followed by Reverse KL. The objective function that achieved the best generalization regard to this evaluation metric is the Generalized I, followed by \textit{MSE}, Forward KL and Cross Entropy, in that order. It is good to point out that \textit{MSE} and Cross Entropy achieved very good result on training data, showing that these model has a very good generalization rate (difference between test and train). The objective function with high overfitting phenomena is the Euclidean, which surprisingly only differs from \textit{MSE} in a constant normalization factor.

The results on the ranking metric, \textit{NDCG}, are shown in Table \ref{ndcg}. Similar to the previous reported results, the worst behave are with Itakura Saito distance and Reverse KL, also happen with the best results (highlight with bold text). Generalized I, Cross Entropy and Jensen-Shannon show the best generalization and performance on test set, also is \textit{MSE}. This results show that the objective function based on probabilities are the best in sort the categories (based on the probabilities), so if this were the objective, this functions are optimal.

\begin{table}[t]
\centering
\begin{tabular}{|c|c|c|c|} \hline
B & Objective function & \textit{train} & \textit{test}  \\ \hline
 & \textit{MSE} & \textbf{83,910} & 47,747 \\ \hline
& \textit{RMSE} & 75,773 & 40,804 \\ \hline
& \textit{Reverse KL} & 38,107 & 29,519\\ \hline
& \textit{Cross Entropy} & 80,075 & 45,352\\ \hline
& \textit{Jensen-Shannon} & 75,486 & 41,543 \\ \hline
$\checkmark$ & \textit{Forward KL} & 79,144 & 46,005 \\ \hline
$\checkmark$ & \textit{Itakura-Saito} & 16,584 & 17,067 \\ \hline
$\checkmark$ & \textit{Generalized I} & 78,944 & \textbf{48,934} \\ \hline
$\checkmark$ & \textit{Squared Euclidean} & 80,333 & 39,156 \\ \hline
\end{tabular}
\caption{Results of percentage \textit{macro F1} metric in both set. $B$ refers to Bregman divergences.}
\label{macrof1}
\end{table}

\begin{table}[t]
\centering
\begin{tabular}{|c|c|c|c|} \hline
B & Objective function & \textit{train} & \textit{test}  \\ \hline
 & \textit{MSE} & 96,847 & \textbf{94,708} \\ \hline
& \textit{RMSE} & 96,823 & 94,559 \\ \hline
& \textit{Reverse KL} & 93,914  & 93,080 \\ \hline
& \textit{Cross Entropy} & \textbf{97,476} & \textbf{94,711} \\ \hline
& \textit{Jensen-Shannon} & \textbf{97,276} & \textbf{94,955} \\ \hline
$\checkmark$ & \textit{Forward KL} & \textbf{97,403} & 94,580\\ \hline
$\checkmark$ & \textit{Itakura-Saito} & 91,867 & 91,805 \\ \hline
$\checkmark$ & \textit{Generalized I} & \textbf{97,364} & \textbf{94,779} \\ \hline
$\checkmark$ & \textit{Squared Euclidean} & 96,824 & 94,462 \\ \hline
\end{tabular}
\caption{Results of percentage \textit{NDCG} metric in both set. $B$ refers to Bregman divergences. Bold represent the four best results in each set.}
\label{ndcg}
\end{table}

As we shown an standard alternative metric to evaluate how the model gives the probabilities of each category, Table \ref{averageranking} measure \textit{Accuracy on Ranking Decrease} (Equation \ref{accuracyranking}). Again some results are repeated, as the worst behave is with Itakura Saito distance and Reverse KL. The objective function with higher score in training set is Forward KL and Jensen-Shannon is the one that generalizes better. Another functions that still has good result are again the based on probabilities: Cross Entropy and Generalized I.

\begin{table}[t]
\centering
\begin{tabular}{|c|c|c|c|} \hline
B & Objective function & \textit{train} & \textit{test}  \\ \hline
 & \textit{MSE} & 28,835 & 16,851 \\ \hline
& \textit{RMSE} & 28,252 & 17,214 \\ \hline
& \textit{Reverse KL} & 20,175 & 15,862 \\ \hline
& \textit{Cross Entropy} & 30,652 & 17,267 \\ \hline
& \textit{Jensen-Shannon} & 30,625 & \textbf{18,657} \\ \hline
$\checkmark$ & \textit{Forward KL} & \textbf{31,007} & 16,788 \\ \hline
$\checkmark$ & \textit{Itakura-Saito} & 11,262 & 11,256 \\ \hline
$\checkmark$ & \textit{Generalized I} & 30,659 & 17,276 \\ \hline
$\checkmark$ & \textit{Squared Euclidean} & 28,350 & 16,700 \\ \hline\end{tabular}
\caption{Results of percentage \textit{Accuracy on ranking decrease} metric in both set. $B$ refers to Bregman divergences. Bold represent the four best results in each set.}
\label{averageranking}
\end{table}

Also the architecture of Figure \ref{arquitectura} was test over the data and the result maintain. The change is that Jensen-Shannon divergence stands over training and test set based on  macro F1 metric. \\

About the results on the text dataset (\textit{Stock tweets emotions}) we obtain similar results so we dont show it here. There is an exception with respect to Reverse KL, that overpass trough all other objective function. This can be because this is a highly unbalanced dataset and this objective function act as a regularized by itself, thanks to maximize the entropy of the prediction and minimize the Cross Entropy between the prediction and real, as you can see in decompose Equation 5. Reverse KL does not have the expected result, of act as an regularize, on GalaxyZoo. \\

Summarizing all the experimented we have that the Bregman divergence functions as a family, that shared the same properties, does not necessary have a good behavior. This could be due to the fact that this are not metrics, and the missing properties of symmetric and triangular inequality can improve the behavior. For example making KL symmetric (Jensen-Shannon divergence) improve some results on the test set.

The commonly used objective function for classification with probabilities, Cross Entropy, turned out to stand in a good way, with good results on different metrics and convergence. While Generalized I, despite not being very studied or chosen in works, present a very good behavior on the metrics and the best generalization.

Despite that Cross Entropy is the same in optimization that Forward KL, except by the entropy of the real probability $H(p)$ that does not depend of the parameters of the model.

\begin{equation}
H(p,q) = H(p) + KL(p||q)
\end{equation}

$H(p)$ turns out zero on the partial derivative. However, this two function reach different values in the optimization, Forward KL stay below of Cross Entropy in the results. Similar case is between \textit{MSE} and Squared Euclidean distance (SSE), in where the difference is only a multiplicative constant, $\frac{1}{K}$, and achieve different results. This could be because the stochastic optimization of the algorithms, as they have the same global minimum, or because the functional derivative of Keras does not have a good precision. 

Here the results show that a slightly change on the objective function of the model can change drastically the results. Since the objective is the same, the curvature of the first derivative is different, amplifying or reducing it.

\section{Conclusion}
In this work report we studied different objective function to optimize a problem of estimate the probabilities of the data and measure various metrics to evaluate quality.

Some results reflect correlation among the functions based on probabilities. As Cross Entropy, Generalized I, Jensen-Shannon and KL show good results on the ranking metrics, it indicates that the models achieved to imitate the order of the category on the data, also the probabilities.

The in-expected result found that the analytically same in optimization objective function achieved different results on all the metrics may be cause different factors. The optimization framework, the stochastic of the optimization algorithm or maybe because the factors that are ignored in optimization may have a contribution.

\bibliographystyle{apalike}
\bibliography{bibliography}

\begin{thebibliography}{}

\bibitem[Banerjee et~al., 2005]{banerjee2005clustering}
Banerjee, A., Merugu, S., Dhillon, I.~S., and Ghosh, J. (2005).
\newblock Clustering with bregman divergences.
\newblock {\em Journal of machine learning research}, 6(Oct):1705--1749.

\bibitem[Bregman, 1967]{bregman1967relaxation}
Bregman, L.~M. (1967).
\newblock The relaxation method of finding the common point of convex sets and
  its application to the solution of problems in convex programming.
\newblock {\em USSR computational mathematics and mathematical physics},
  7(3):200--217.

\bibitem[Cao et~al., 2007]{cao2007learning}
Cao, Z., Qin, T., Liu, T.-Y., Tsai, M.-F., and Li, H. (2007).
\newblock Learning to rank: from pairwise approach to listwise approach.
\newblock In {\em Proceedings of the 24th international conference on Machine
  learning}, pages 129--136. ACM.

\bibitem[Chen et~al., 2008]{chen2008metrics}
Chen, P., Chen, Y., Rao, M., et~al. (2008).
\newblock Metrics defined by bregman divergences: Part 2.
\newblock {\em Communications in Mathematical Sciences}, 6(4):927--948.

\bibitem[Chollet et~al., 2015]{chollet2015keras}
Chollet, F. et~al. (2015).
\newblock Keras.

\bibitem[Chung et~al., 2014]{chung2014empirical}
Chung, J., Gulcehre, C., Cho, K., and Bengio, Y. (2014).
\newblock Empirical evaluation of gated recurrent neural networks on sequence
  modeling.
\newblock {\em arXiv preprint arXiv:1412.3555}.

\bibitem[Glorot and Bengio, 2010]{glorot2010understanding}
Glorot, X. and Bengio, Y. (2010).
\newblock Understanding the difficulty of training deep feedforward neural
  networks.
\newblock In {\em Proceedings of the thirteenth international conference on
  artificial intelligence and statistics}, pages 249--256.

\bibitem[Kingma and Ba, 2014]{kingma2014adam}
Kingma, D.~P. and Ba, J. (2014).
\newblock Adam: A method for stochastic optimization.
\newblock {\em arXiv preprint arXiv:1412.6980}.

\bibitem[LeCun et~al., 1995]{lecun1995convolutional}
LeCun, Y., Bengio, Y., et~al. (1995).
\newblock Convolutional networks for images, speech, and time series.
\newblock {\em The handbook of brain theory and neural networks},
  3361(10):1995.

\bibitem[LeCun et~al., 2015]{lecun2015deep}
LeCun, Y., Bengio, Y., and Hinton, G. (2015).
\newblock Deep learning.
\newblock {\em nature}, 521(7553):436.

\bibitem[Lin, 1991]{lin1991divergence}
Lin, J. (1991).
\newblock Divergence measures based on the shannon entropy.
\newblock {\em IEEE Transactions on Information theory}, 37(1):145--151.

\bibitem[Lintott et~al., 2008]{lintott2008galaxy}
Lintott, C.~J., Schawinski, K., Slosar, A., Land, K., Bamford, S., Thomas, D.,
  Raddick, M.~J., Nichol, R.~C., Szalay, A., Andreescu, D., et~al. (2008).
\newblock Galaxy zoo: morphologies derived from visual inspection of galaxies
  from the sloan digital sky survey.
\newblock {\em Monthly Notices of the Royal Astronomical Society},
  389(3):1179--1189.

\bibitem[Nowozin et~al., 2016]{nowozin2016f}
Nowozin, S., Cseke, B., and Tomioka, R. (2016).
\newblock f-gan: Training generative neural samplers using variational
  divergence minimization.
\newblock In {\em Advances in Neural Information Processing Systems}, pages
  271--279.

\bibitem[Snow et~al., 2008]{snow2008cheap}
Snow, R., O'Connor, B., Jurafsky, D., and Ng, A.~Y. (2008).
\newblock Cheap and fast---but is it good?: evaluating non-expert annotations
  for natural language tasks.
\newblock In {\em Proceedings of the conference on empirical methods in natural
  language processing}, pages 254--263. Association for Computational
  Linguistics.

\bibitem[Sugiyama et~al., 2012]{sugiyama2012density}
Sugiyama, M., Suzuki, T., and Kanamori, T. (2012).
\newblock Density-ratio matching under the bregman divergence: a unified
  framework of density-ratio estimation.
\newblock {\em Annals of the Institute of Statistical Mathematics},
  64(5):1009--1044.

\bibitem[Thomas, 1991]{thomas1991elements}
Thomas, J. (1991).
\newblock Elements of information theory.

\bibitem[Vemuri et~al., 2011]{vemuri2011total}
Vemuri, B.~C., Liu, M., Amari, S.-I., and Nielsen, F. (2011).
\newblock Total bregman divergence and its applications to dti analysis.
\newblock {\em IEEE Transactions on medical imaging}, 30(2):475--483.

\end{thebibliography}

\end{document}